%% file: 0_main_TSformer.tex
\documentclass{article}
\usepackage{ijcai21}

\usepackage{graphicx} 

\usepackage{subfigure}
\usepackage{times}
\usepackage{soul}
\usepackage{url}
\usepackage[hidelinks]{hyperref}
\usepackage[utf8]{inputenc}
\usepackage[small]{caption}
\usepackage{amsmath}
\usepackage{amsthm}
\usepackage{booktabs}
\usepackage{algorithm}
\usepackage{algorithmic}
\usepackage{multirow}
\usepackage{xcolor}
\usepackage[hang]{footmisc}
\title{Transformers in Time Series: A Survey}

\author{Qingsong Wen\textsuperscript{\rm 1}, Tian Zhou\textsuperscript{\rm 2}, Chaoli Zhang\textsuperscript{\rm 2}, Weiqi Chen\textsuperscript{\rm 2}, Ziqing Ma\textsuperscript{\rm 2}, Junchi Yan\textsuperscript{\rm 3}, Liang Sun\textsuperscript{\rm 1}
    \affiliations
    \textsuperscript{\rm 1}DAMO Academy, Alibaba Group, Bellevue, USA\\
    \textsuperscript{\rm 2}DAMO Academy, Alibaba Group, Hangzhou, China\\
    \textsuperscript{\rm 3}Department of CSE, MoE Key Lab of Artificial Intelligence, Shanghai Jiao Tong University
    \emails{
    \{qingsong.wen, tian.zt, chaoli.zcl, jarvus.cwq, maziqing.mzq, liang.sun\}@alibaba-inc.com}, yanjunchi@sjtu.edu.cn
}

\begin{document}

\maketitle
 
\begin{abstract}
Transformers have achieved superior performances in many tasks in natural language processing and computer vision, which also triggered great interest in the time series community. Among multiple advantages of Transformers, the ability to capture long-range dependencies and interactions is especially attractive for time series modeling, leading to exciting progress in various time series applications. In this paper, we systematically review Transformer schemes for time series modeling by highlighting their strengths as well as limitations. In particular, we examine the development of time series Transformers in two perspectives. From the perspective of network structure, we summarize the 
adaptations and modifications that have been made to Transformers in order to accommodate the challenges in time series analysis. From the perspective of applications, we categorize time series Transformers based on common tasks including forecasting, anomaly detection, and classification. Empirically, we perform robust analysis, model size analysis, and seasonal-trend decomposition analysis to study how Transformers perform in time series. Finally, we discuss and suggest future directions to provide useful research guidance. A corresponding resource that has been continuously updated can be found in the GitHub repository\footnote{\url{https://github.com/qingsongedu/time-series-transformers-review}}.
\end{abstract}

\input{1_intro.tex}

\input{1_preliminaries.tex}


\input{2_taxonomy.tex}

\input{3_0_Xformer_modify_for_TS.tex}

\section{Applications of Time Series Transformers}\label{secapp}
In this section, we review the applications of Transformer to important time series tasks, including forecasting, anomaly detection, and classification.
\input{3_1_forecasting}
\input{3_2_Anomaly}

\input{3_3_classification_cluster}

\input{5_1_evaluation_discuss.tex}

\input{5_2_research_direction.tex}

\input{6_conclusion.tex}


%
\bibliographystyle{named}
{\footnotesize
\bibliography{7_reference}}



\end{document}

%% file: 1_intro.tex
\section{Introduction}

The innovation of Transformer in deep learning~\cite{vaswani2017attention} has brought great interests recently due to its excellent performances in natural language processing (NLP)~\cite{kenton2019bert}, computer vision (CV)~\cite{Dosovitskiy2020AnII}, and speech processing~\cite{dong2018speech}. Over the past few years, numerous Transformers have been proposed to advance the state-of-the-art performances of various tasks significantly. There are quite a few literature reviews from different aspects, such as in NLP applications~\cite{han2021pre}, CV applications~\cite{han2020survey}, and efficient Transformers~\cite{tay2020efficient}.


Transformers have shown great modeling ability for long-range dependencies and interactions in sequential data and thus are appealing to time series modeling. 
Many variants of Transformer have been proposed to address special challenges in time series modeling and have been successfully applied to various time series tasks, such as forecasting~\cite{li2019enhancing,zhou2022fedformer}, anomaly detection~\cite{xu2022anomalyTrans,tuli2022tranad}, and classification~\cite{zerveas2021transformer,yang2021voice2series}. Specifically, seasonality or periodicity is an important feature of time series~\cite{WenRobustPeriod20}. How to effectively model long-range and short-range temporal dependency and capture seasonality simultaneously remains a challenge~\cite{xu2021autoformer,wen2022robust}. 
We note that there exist several surveys related to deep learning for time series, including forecasting~\cite{tsDLSurvey21,benidis2020neural,DLTSSurvey21}, classification~\cite{ismail2019deep}, anomaly detection~\cite{choi2021deep,blazquez2021review}, and data augmentation~\cite{wentsaug2021}, but there is no comprehensive survey for Transformers in time series. As Transformer for time series is an emerging subject in deep learning, a systematic and comprehensive survey on time series Transformers would greatly benefit the time series community.

In this paper, we aim to fill the gap by
summarizing the main developments of time series Transformers. We first give a brief introduction about vanilla Transformer, and then propose a new taxonomy from perspectives of both network modifications and application domains for time series Transformers. 
For network modifications, we discuss the improvements made on both low-level (i.e. module) and high-level (i.e. architecture) of Transformers, with the aim to optimize the performance of time series modeling.
For applications, we analyze and summarize Transformers for popular time series tasks including forecasting, anomaly detection, and classification. 
For each time series Transformer, we analyze its insights, strengths, and limitations. 
To provide practical guidelines on how to effectively use Transformers for time series modeling, we conduct extensive empirical studies that examine multiple aspects of time series modeling, including robustness analysis, model size analysis, and seasonal-trend decomposition analysis.
We conclude this work by discussing possible future directions for time series Transformers, including inductive biases for time series Transformers, Transformers and GNN for time series,
pre-trained Transformers for time series, Transformers with architecture level variants, and Transformers with NAS for time series. To the best of our knowledge, this is the first work to comprehensively and systematically review the key developments of Transformers for modeling time series data. We hope this survey will ignite further research interests in time series Transformers.



%% file: 1_preliminaries.tex
\section{Preliminaries of the Transformer}

\subsection{Vanilla Transformer}
The vanilla Transformer~\cite{vaswani2017attention} follows most competitive neural sequence models with an encoder-decoder structure. Both encoder and decoder are composed of multiple identical blocks. Each encoder block consists of a multi-head self-attention module and a position-wise feed-forward network while each decoder block inserts cross-attention models between the multi-head self-attention module and the position-wise feed-forward network. 

\subsection{Input Encoding and Positional Encoding}
Unlike LSTM or RNN, the vanilla Transformer has no recurrence. Instead, it utilizes the positional encoding added in the input embeddings, to model the sequence information. We summarize some positional encodings below. 

\subsubsection{Absolute Positional Encoding}\label{subsubsec:vallina positional encoding}
In vanilla Transformer, for each position index $t$, encoding vector is given by
\begin{equation}
{PE}(t)_i = 
\left\{ 
\begin{aligned} 
\sin(\omega_{i}t) \quad i\%2=0\\
\cos(\omega_{i}t) \quad i\%2=1
\end{aligned}
\right.
\end{equation}
where $\omega_i$ is the hand-crafted frequency for each dimension.
Another way is to learn a set of positional embeddings for each position which is more flexible~\cite{kenton2019bert,gehring2017convolutional}. 

\subsubsection{Relative Positional Encoding}
Following the intuition that pairwise positional relationships between input elements is more beneficial than positions of elements, relative positional encoding methods have been proposed. For example, one of such methods is to add a learnable relative positional embedding to keys of attention mechanism~\cite{shaw2018self}.

Besides the absolute and relative positional encodings, there are methods using hybrid positional encodings that combine them together~\cite{ke2020rethinking}. Generally, the positional encoding is added to the token embedding and fed to Transformer.

\subsection{Multi-head Attention}
With Query-Key-Value (QKV) model, the scaled dot-product attention used by Transformer is given by 
\begin{equation}
    Attention(\mathbf{Q,K,V})=softmax(\frac{\mathbf{QK^T}}{\sqrt{D_k}})\mathbf{V}
\end{equation} 
where queries $\mathbf{Q}\in\mathcal{R}^{N\times D_k}$, keys $\mathbf{K}\in\mathcal{R}^{M\times D_k}$, values $\mathbf{V}\in\mathcal{R}^{M\times D_v}$, $N, M$ denote the lengths of queries and keys (or values), and $D_k, D_v$ denote the dimensions of keys (or queries) and values.
Transformer uses multi-head attention with $H$ different sets of learned projections instead of a single attention function as
\begin{equation} \notag
\resizebox{1\hsize}{!}{$
    \operatorname{MultiHeadAttn}(\mathbf{Q,K,V}) = \operatorname{Concat}{(head_1, {\cdots}, head_H)}\mathbf{W}^{O},
    $}
\end{equation}
where 
$
    head_i = Attention(\mathbf{QW}^{Q}_{i}, \mathbf{KW}^{K}_{i}, \mathbf{VW}^{V}_{i}).
$

\subsection{Feed-forward and Residual Network}
The feed-forward network is a fully connected module as
\begin{equation}
    {FFN}(\mathbf{H}')={ReLU}(\mathbf{H}{'}\mathbf{W}^{1}+\mathbf{b}^{1})\mathbf{W}^{2}+\mathbf{b}^{2},
\end{equation}
where $\mathbf{H}'$ is outputs of previous layer, $\mathbf{W}^{1} \in \mathcal{R}^{D_m\times D_f}$, $\mathbf{W}^{2}\in \mathcal{R}^{D_f\times D_m}$,  $\mathbf{b}^{1}\in \mathcal{R}^{D_f}$, $\mathbf{b}^{2} \in \mathcal{R}^{D_m}$ are trainable parameters.
In a deeper module, a residual connection module followed by a layer normalization module is inserted around each module. That is, 
\begin{align}
\mathbf{H}{'} &=  \operatorname{LayerNorm}(SelfAttn(\mathbf{X})+\mathbf{X}),\\
\mathbf{H} &= \operatorname{LayerNorm}(FFN(\mathbf{H}{'})+\mathbf{H}{'}),
\end{align}
where $SelfAttn{(.)}$ denotes self-attention module and $LayerNorm(.)$ denotes the layer normalization operation.

%% file: 2_taxonomy.tex
\begin{figure}[!t]
\centering
    \includegraphics[width=0.48\textwidth]{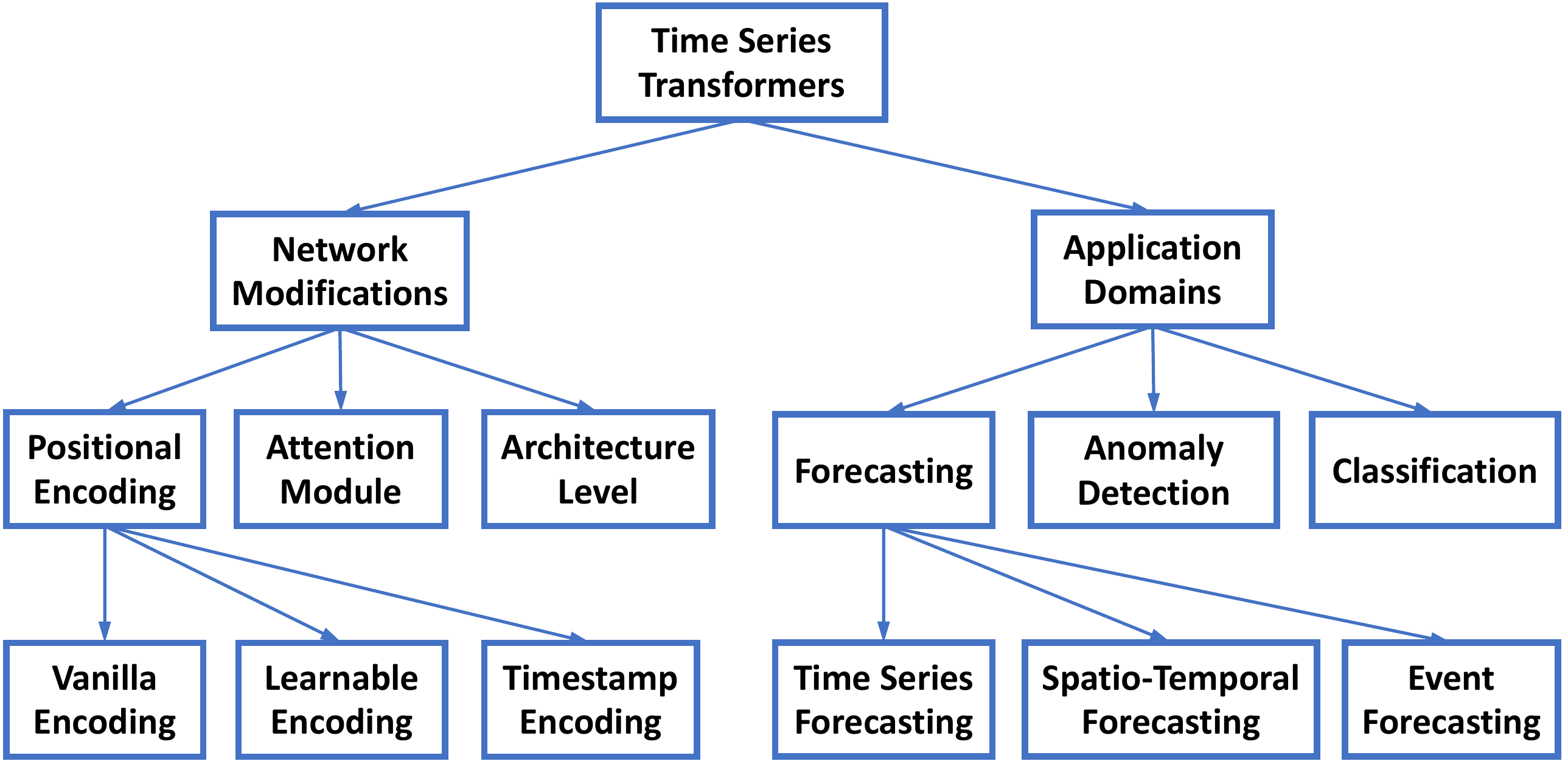}
    \vspace{-4mm}
    \caption{Taxonomy of Transformers for time series modeling from the perspectives of network modifications and application domains.}
    \vspace{-2mm}
\label{fig:ts_xformers}
\end{figure}
\section{Taxonomy of Transformers in Time Series}
To summarize the existing time series Transformers, we propose a taxonomy from perspectives of network modifications and application domains as illustrated in Fig.~\ref{fig:ts_xformers}. Based on the taxonomy, we review the existing time series Transformers systematically. 
From the perspective of network modifications, we summarize the changes made on both module level and architecture level of Transformer in order to accommodate special challenges in time series modeling. 
From the perspective of applications, we classify time series Transformers based on their application tasks, including forecasting, anomaly detection, and classification. 
In the following two sections, we would delve into the existing time series Transformers from these two perspectives.

%% file: 3_0_Xformer_modify_for_TS.tex
\section{Network Modifications for Time Series}

\subsection{Positional Encoding}
As the ordering of time series matters, it is of great importance to encode the positions of input time series into Transformers. A common design is to first encode positional information as vectors and then inject them into the model as an additional input together with the input time series. How to obtain these vectors when modeling time series with Transformers can be divided into three main categories.


\textbf{Vanilla Positional Encoding.}\quad A few works \cite{li2019enhancing} simply introduce vanilla positional encoding (Section \ref{subsubsec:vallina positional encoding}) used in  \cite{vaswani2017attention}, which is then added to the input time series embeddings and fed to Transformer. Although this approach can extract some positional information from time series, they were unable to fully exploit the important features of time series data. 


\textbf{Learnable Positional Encoding.}\quad As the vanilla positional encoding is hand-crafted and less expressive and adaptive, several studies found that learning appropriate positional embeddings from time series data can be much more effective. Compared to fixed vanilla positional encoding, learned embeddings are more flexible and can adapt to specific tasks. ~\cite{zerveas2021transformer} introduces an embedding layer in Transformer that learns embedding vectors for each position index jointly with other model parameters. \cite{lim2021temporal} uses an LSTM network to encode positional embeddings, which can better exploit sequential ordering information in time series.

\textbf{Timestamp Encoding.}\quad When modeling time series in real-world scenarios, the timestamp information is commonly accessible, including calendar timestamps (e.g., second, minute, hour, week, month, and year) and special timestamps (e.g., holidays and events). These timestamps are quite informative in real applications but hardly leveraged in vanilla Transformers. To mitigate the issue, Informer~\cite{zhou2021informer} proposed to encode timestamps as additional positional encoding by using learnable embedding layers. A similar timestamp encoding scheme was used in Autoformer \cite{xu2021autoformer} and FEDformer~\cite{zhou2022fedformer}. 









\subsection{Attention Module}
Central to Transformer is the self-attention module. It can be viewed as a fully connected layer with weights that are dynamically generated based on the pairwise similarity of input patterns. As a result, it shares the same maximum path length as fully connected layers, but with a much less number of parameters, making it suitable for modeling long-term dependencies. 

As we show in the previous section the self-attention module in the vanilla Transformer has a time and memory complexity of $\mathcal{O}(N^2)$ ($N$ is the input time series length), which becomes the computational bottleneck when dealing with long sequences. Many efficient Transformers were proposed to reduce the quadratic complexity that can be classified into two main categories: 
(1) explicitly introducing a sparsity bias into the attention mechanism like LogTrans~\cite{li2019enhancing} and Pyraformer~\cite{liu2022pyraformer}; (2) exploring the low-rank property of the self-attention matrix to speed up the computation, e.g. Informer~\cite{zhou2021informer} and FEDformer~\cite{zhou2022fedformer}. Table~\ref{tab:compare_complexity} shows both the time and memory complexity of popular Transformers applied to time series modeling, and more details about these models will be discussed in Section~\ref{secapp}.

\input{tables/complexity}

\subsection{Architecture-based Attention Innovation}
To accommodate individual modules in Transformers for modeling time series, a number of works ~\cite{zhou2021informer,liu2022pyraformer} seek to renovate Transformers on the architecture level. Recent works introduce hierarchical architecture into Transformer to take into account  the multi-resolution aspect of time series. Informer~\cite{zhou2021informer} inserts max-pooling layers with stride 2 between attention blocks, which down-sample series into its half slice. Pyraformer~\cite{liu2022pyraformer} designs a $C$-ary tree-based attention mechanism, in which nodes at the finest scale correspond to the original time series, while nodes in the coarser scales represent series at lower resolutions. Pyraformer developed both intra-scale and inter-scale attentions in order to better capture temporal dependencies across different resolutions. Besides the ability to integrate information at different multi-resolutions, a hierarchical architecture also enjoys the benefits of efficient computation, particularly for long-time series.

%% file: tables/complexity.tex

\begin{table}
\centering
\caption{Complexity comparisons of popular time series Transformers with different attention modules.} 
\vspace{-3mm}
\scalebox{0.75}{
\begin{tabular}{l|ccc}
\hline \multirow{2}{*}{ Methods } & \multicolumn{2}{|c}{ Training } & Testing \\
\cline { 2 - 4 } & Time & Memory & Steps \\
\hline 

Transformer~\cite{vaswani2017attention} & $\mathcal{O}\left(N^{2}\right)$ & $\mathcal{O}\left(N^{2}\right)$ & $N$ \\

LogTrans~\cite{li2019enhancing} & $\mathcal{O}(N \log N)$ & $\mathcal{O}\left(N \log N\right)$ & 1 \\

Informer~\cite{zhou2021informer} & $\mathcal{O}(N \log N)$ & $\mathcal{O}(N \log N)$ & 1 \\

Autoformer~\cite{xu2021autoformer} & $\mathcal{O}(N \log N)$ & $\mathcal{O}(N \log N)$ & 1 \\

Pyraformer~\cite{liu2022pyraformer} & $\mathcal{O}(N)$ & $\mathcal{O}(N)$ & 1 \\

Quatformer~\cite{chen2022quatformer}& $\mathcal{O}(2cN)$ & $\mathcal{O}(2cN)$ & 1 \\

FEDformer~\cite{zhou2022fedformer} & $\mathcal{O}(N)$ & $\mathcal{O}(N)$ & 1 \\

Crossformer~\cite{zhang2023cross} & $\mathcal{O}(\frac{D}{L_{seg}^2}N^2)$ & $\mathcal{O}(N)$ & 1 \\

\hline
\end{tabular}
}
\label{tab:compare_complexity}
\vspace{-3mm}
\end{table}

%% file: 3_1_forecasting.tex
\subsection{Transformers in Forecasting}
Here we examine three common types of forecasting tasks here, i.e. time series forecasting, spatial-temporal forecasting, and event forecasting. 

\subsubsection{Time Series Forecasting}


A lot of work has been done to design new Transformer variants for time series forecasting tasks in the latest years. Module-level and architecture-level variants are two large categories and the former consists of the majority of the up-to-date works. 

\begin{figure}[!t]
\centering
    \includegraphics[width=0.42\textwidth]{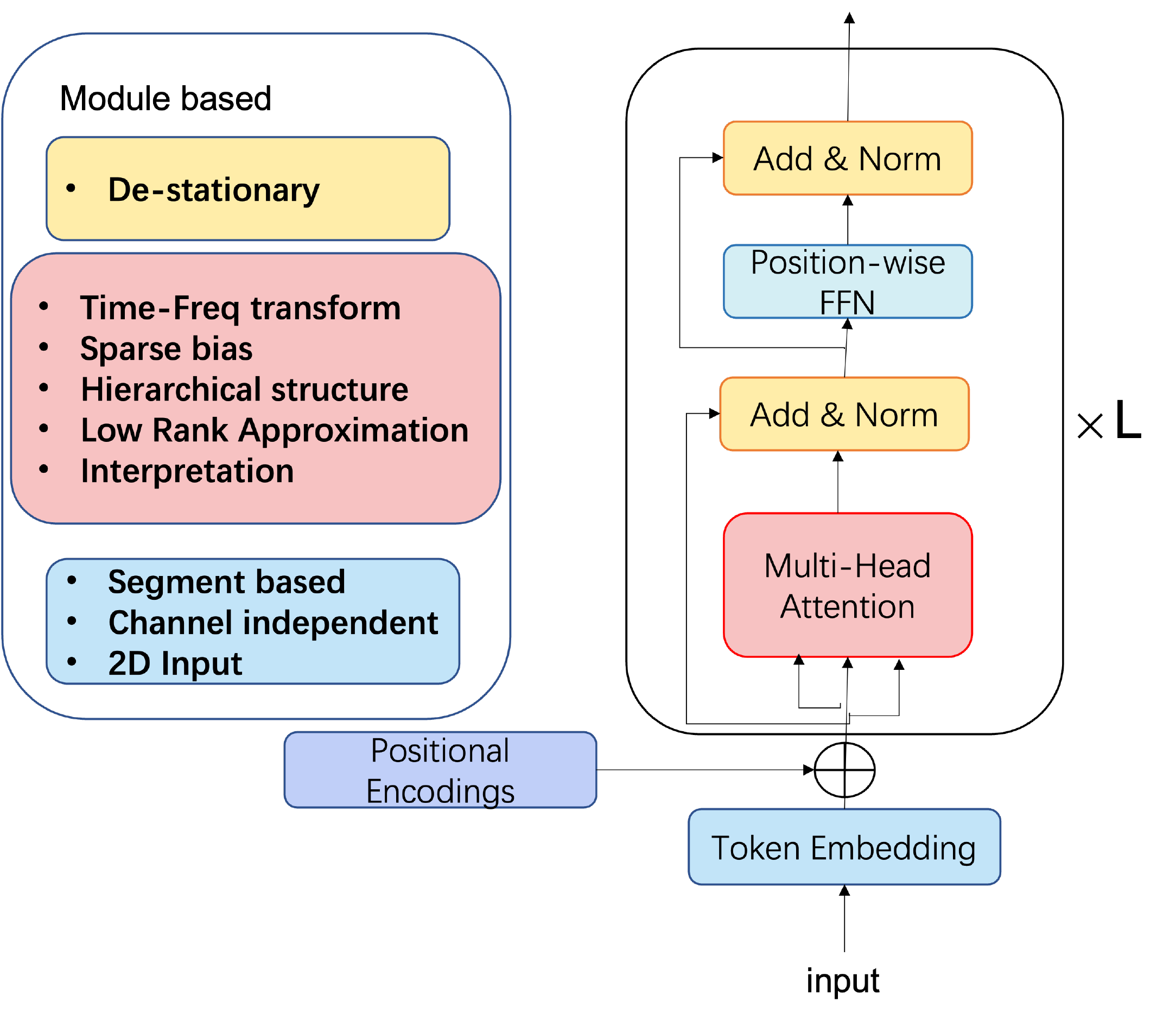}\vspace{-4mm}
    \caption{Categorization of module-level Transformer variants for time series forecasting.}\vspace{-4mm}
    \label{fig:Categorization}
\end{figure}

\paragraph{Module-level variants}
In the module-level variants for time series forecasting, their main architectures are similar to the vanilla Transformer with minor changes. Researchers introduce various time series inductive biases to design new modules. The following summarized work consists of three different types: designing new attention modules, exploring the innovative way to normalize time series data, and utilizing the bias for token inputs, as shown in Figure \ref{fig:Categorization}. 

The first type of variant for module-level Transformers is to design new attention modules, which is the category with the largest proportion. 
Here we first describe six typical works: LogTrans~\cite{li2019enhancing}, Informer~\cite{zhou2021informer}, AST~\cite{wu2020adversarial}, Pyraformer~\cite{liu2022pyraformer}, Quatformer~\cite{chen2022quatformer}, and FEDformer~\cite{zhou2022fedformer}, all of which exploit sparsity inductive bias or low-rank approximation to remove noise and achieve a low-order calculation complexity. 
LogTrans~\cite{li2019enhancing} proposes convolutional self-attention by employing causal convolutions to generate queries and
keys in the self-attention layer. It introduces sparse bias, a Logsparse mask, in self-attention model that reduces computational complexity from $\mathcal{O}(N^2)$ to $\mathcal{O}(N\log N)$.
Instead of using explicit sparse bias, Informer~\cite{zhou2021informer} selects dominant queries based on queries and key similarities, thus achieving similar improvements as LogTrans in computational complexity. It also designs a generative style decoder to produce long-term forecasting directly and thus avoids accumulative error in using one forward-step prediction for long-term forecasting.
AST~\cite{wu2020adversarial} uses a generative adversarial encoder-decoder framework to train a sparse Transformer model for time series forecasting. It shows that adversarial training can improve time series forecasting by directly shaping the output distribution of the network to avoid error accumulation through one-step ahead inference. Pyraformer~\cite{liu2022pyraformer} designs a hierarchical pyramidal attention module with a binary tree following the path, to capture temporal dependencies of different ranges with linear time and memory complexity. 
FEDformer~\cite{zhou2022fedformer} applies attention operation in the frequency domain with Fourier transform and wavelet transform. It achieves a linear complexity by randomly selecting a fixed-size subset of frequency. Note that due to the success of Autoformer and FEDformer, it has attracted more attention in the community to explore self-attention mechanisms in the frequency domain for time series modeling. Quatformer~\cite{chen2022quatformer} proposes learning-to-rotate attention (LRA) based on quaternions that introduce learnable period and phase information to depict intricate periodical patterns. Moreover, it decouples LRA using a global memory to achieve linear complexity. 

The following three works focus on building an explicit interpretation ability of models, which follows the trend of Explainable Artificial Intelligence (XAI). TFT~\cite{lim2021temporal} designs a multi-horizon forecasting model with static covariate encoders, gating feature selection, and temporal self-attention decoder. It encodes and selects useful information from various covariates to perform forecasting. It also preserves interpretability by incorporating global, temporal dependency, and events. 
ProTran~\cite{tang2021probabilistic} and SSDNet~\cite{lin2021ssdnet} combine
Transformer with state space models to provide probabilistic forecasts. 
ProTran designs a generative modeling and inference procedure based on variational inference. SSDNet first uses Transformer to learn the temporal pattern and estimate the parameters of SSM, and then applies SSM to perform the seasonal-trend decomposition and maintain the interpretable ability.

The second type of variant for module-level Transformers is the way to normalize time series data. To the best of our knowledge, Non-stationary Transformer~\cite{liu2022non} is the only work that mainly focuses on modifying the normalization mechanism as shown in Figure~\ref{fig:Categorization}. It explores the over-stationarization problem in time series forecasting tasks with a relatively simple plugin series stationary and De-stationary module to modify and boost the performance of various attention blocks. 

The third type of variant for module-level Transformer is utilizing the bias for token input. 
Autoformer~\cite{xu2021autoformer} adopts a segmentation-based representation mechanism. 
It devises a simple seasonal-trend decomposition architecture with an auto-correlation mechanism working as an attention module. The auto-correlation block measures the time-delay similarity between inputs signal and aggregates the top-k similar sub-series to produce the output with reduced complexity.
PatchTST~\cite{Nie2022ATS} utilizes channel-independent where each channel contains a single univariate time series that shares the same embedding within all the series, and subseries-level patch design which segmentation of time series into subseries-level patches that are served as input tokens to Transformer. Such ViT \cite{Dosovitskiy2020AnII} alike design improves its numerical performance in long-time time-series forecasting tasks a lot. 
Crossformer~\cite{zhang2023cross} proposes a Transformer-based model utilizing cross-dimension dependency for multivariate time series forecasting. The input is embedded into a 2D vector array through the novel dimension-segment-wise embedding to preserve time and dimension information. Then, a two-stage attention layer is used to efficiently capture the cross-time and cross-dimension dependency.  

\paragraph{Architecture-level variants}
Some works start to design a new transformer architecture beyond the scope of the vanilla transformer. 
Triformer~\cite{cirstea2022triformer} design a triangular,variable-specific patch attention. It has a triangular tree-type structure as the later input size shrinks exponentially and a set of variable-specific parameters making a multi-layer Triformer maintain a lightweight and linear complexity.
Scaleformer~\cite{shabani2023scaleformer} proposes a multi-scale framework that can be applied to the baseline transformer-based time series forecasting models (FEDformer\cite{zhou2022fedformer}, Autoformer\cite{xu2021autoformer}, etc.). It can improve the baseline model's performance by iteratively refining the forecasted time series at multiple scales with shared weights.  
\paragraph{Remarks}
Note that DLinear~\cite{zeng2022transformers} questions the necessity of using Transformers for long-term time series forecasting, and shows that a simpler MLP-based model can achieve better results compared to some Transformer baselines through empirical studies. 
However, we notice that a recent Transformer model PatchTST \cite{Nie2022ATS} achieves a better numerical result compared to DLinear for long-term time series forecasting. 
Moreover, there is a thorough theoretical study \cite{Yun2019AreTU} showing that the Transformer models are universal approximators of sequence-to-sequence functions.
It is a overclaim to question the potential of any type of method for time series forecasting based solely on experimental results from some variant instantiations of such method, especially for Transformer models which already demonstrate the performances in most machine learning-based tasks. Therefore, we conclude that summarizing the recent Transformer-based models for time series forecasting is necessary and would benefit the whole community.



\subsubsection{Spatio-Temporal Forecasting}
In spatio-temporal forecasting, both temporal and spatio-temporal dependencies are taken into account in time series Transformers for accurate forecasting. 

Traffic Transformer~\cite{cai2020traffic} designs an encoder-decoder structure using a self-attention module to capture temporal-temporal dependencies and a graph neural network module to capture spatial dependencies. Spatial-temporal Transformer~\cite{xu2020spatial} for traffic flow forecasting takes a step further. Besides introducing a temporal Transformer block to capture temporal dependencies, it also designs a spatial Transformer block, together with a graph convolution network, to better capture spatial-spatial dependencies. Spatio-temporal graph Transformer~\cite{yu2020spatio} designs an attention-based graph convolution mechanism that is able to learn a complicated temporal-spatial attention pattern to improve pedestrian trajectory prediction.
Earthformer~\cite{gaoearthformer2022} proposes a cuboid attention for efficient space-time modeling, which decomposes the data into cuboids and applies cuboid-level self-attention in parallel. It shows that Earthformer achieves superior performance in weather and climate forecasting. 
Recently, AirFormer~\cite{liang2023airformer} devises a dartboard spatial self-attention module and a causal temporal self-attention module to efficiently capture spatial correlations and temporal dependencies, respectively. Furthermore, it enhances Transformers with latent variables to capture data uncertainty and improve air quality forecasting.

\subsubsection{Event Forecasting}
Event sequence data with irregular and asynchronous timestamps are naturally observed in many real-life applications, which is in contrast to regular time series data with equal sampling intervals. 
Event forecasting or prediction aims to predict the times and marks of future events given the history of past events, and it is often modeled by temporal point processes (TPP)~\cite{yan2019modeling,shchur2021neural}. 

Recently, several neural TPP models incorporate Transformers in order to improve the performance of event prediction.
Self-attentive Hawkes process (SAHP)~\cite{zhang2020self} and Transformer Hawkes process (THP)~\cite{zuo2020transformer} adopt Transformer encoder architecture to summarize the influence of historical events and compute the intensity function for event prediction. They modify the positional encoding by translating time intervals into sinusoidal functions such that the intervals between events can be utilized. Later, a more flexible named attentive neural datalog through time (A-NDTT)~\cite{mei2022transformer} is proposed to extend SAHP/THP schemes by embedding all possible events and times with attention as well. Experiments show that it can better capture sophisticated event dependencies than existing methods.






%% file: 3_2_anomaly.tex
\subsection{Transformers in Anomaly Detection}

Transformer based architecture also benefits the time series anomaly detection task with the ability to model temporal dependency, which brings high detection quality~\cite{xu2022anomalyTrans}. Besides, 
in multiple studies, including TranAD~\cite{tuli2022tranad}, MT-RVAE~\cite{wang2022variational}, and TransAnomaly~\cite{zhang2021unsupervised}, researchers proposed to combine Transformer with neural generative models, such as VAEs~\cite{kingma2013auto} and GANs~\cite{goodfellow2014generative}, for better performance in anomaly detection. We will elaborate on these models in the following part. 

TranAD~\cite{tuli2022tranad} proposes an adversarial training procedure to amplify reconstruction errors as a simple Transformer-based network tends to miss small deviation of anomaly. GAN style adversarial training procedure is designed by two Transformer encoders and two Transformer decoders to gain stability. Ablation study shows that, if Transformer-based encoder-decoder is replaced, F1 score drops nearly 11\%, indicating the effect of Transformer architecture on time series anomaly detection.

MT-RVAE~\cite{wang2022variational} and TransAnomaly~\cite{zhang2021unsupervised} combine VAE with Transformer, but they share different purposes. TransAnomaly combines VAE with Transformer to allow more parallelization and reduce training costs by nearly 80\%. In MT-RVAE, a multiscale Transformer is designed to extract and integrate time-series information at different scales. It overcomes the shortcomings of traditional Transformers where only local information is extracted for sequential analysis.

GTA~\cite{chen2021learning} combines Transformer with graph-based learning architecture for multivariate time series anomaly detection. 
Note that, MT-RVAE is also for multivariate time series but with few dimensions or insufficient close relationships among sequences where the graph neural network model does not work well. To deal with such challenge, MT-RVAE modifies the positional encoding module and introduces feature-learning module. Instead, GTA contains a graph convolution structure to model the influence propagation process. Similar to MT-RVAE, GTA also considers ``global'' information, yet by replacing vanilla multi-head attention with a multi-branch attention mechanism, that is, a combination of global-learned attention, vanilla multi-head attention, and neighborhood convolution.

AnomalyTrans~\cite{xu2022anomalyTrans} combines Transformer and Gaussian prior-Association to make anomalies more distinguishable. Sharing similar motivation as TranAD, AnomalyTrans achieves the goal in a different way. The insight is that it is harder for anomalies to build strong associations with the whole series while easier with adjacent time points compared with normality. In AnomalyTrans, prior-association and series-association are modeled simultaneously. Besides reconstruction loss, the anomaly model is optimized by the minimax strategy to constrain the prior- and series- associations for more distinguishable association discrepancy.






%% file: 3_3_classification_cluster.tex
\subsection{Transformers in Classification}

Transformer is proved to be effective in various time series classification tasks due to its prominent capability in capturing long-term dependency. 
GTN~\cite{liu2021gated} uses a two-tower Transformer with each tower respectively working on time-step-wise attention and channel-wise attention. To merge the feature of the two towers, a learnable weighted concatenation (also known as `gating') is used. The proposed extension of Transformer achieves state-of-the-art results on 13 multivariate time series classifications. \cite{attention-satellite-classi/rubwurm/2020} studied the self-attention based Transformer for raw optical satellite time series classification and obtained the best results compared with recurrent and convolutional neural networks.
Recently, TARNet~\cite{chowdhury2022tarnet} designs Transformers to learn task-aware data reconstruction that augments classification performance, which utilizes attention score for important timestamps masking and reconstruction and brings superior performance.

Pre-trained Transformers are also investigated in classification tasks. \cite{yuan2020self} studies the Transformer for raw optical satellite image time series classification. The authors use self-supervised pre-trained schema because of limited labeled data. \cite{zerveas2021transformer} introduced an unsupervised pre-trained framework and the model is pre-trained with proportionally masked data. The pre-trained models are then fine-tuned in downstream tasks such as classification. 
\cite{yang2021voice2series} proposes to use large-scale pre-trained speech processing model for downstream time series classification problems and generates 19 competitive results on 30 popular time series classification datasets.




%% file: 5_1_evaluation_discuss.tex
\section{Experimental Evaluation and Discussion}
We conduct preliminary empirical studies on a typical challenging benchmark dataset ETTm2~\cite{zhou2021informer} to analyze how Transformers work on time series data. Since classic statistical ARIMA/ETS~\cite{hyndman2008automatic} models and basic RNN/CNN models perform inferior to Transformers in this dataset as shown in ~\cite{zhou2021informer,xu2021autoformer}, we focus on 
popular time series Transformers with different configurations in the experiments.

\subsubsection{Robustness Analysis}
A lot of works we describe above carefully design attention modules to lower the quadratic
calculation and memory complexity, though they practically use a short fixed-size input to achieve the best result in their reported experiments. It makes us question the actual usage of such an efficient design. We perform a robust experiment with prolonging input sequence length to verify their prediction power and robustness when dealing with long-term input sequences in Table \ref{tab:robustness}. 

\input{tables/robustness}

As in Table \ref{tab:robustness}, when we compare the prediction results with prolonging input length, various Transformer-based model deteriorates quickly. This phenomenon makes a lot of carefully designed Transformers impractical in long-term forecasting tasks since they cannot effectively utilize long input information. More works and designs need to be investigated to fully utilize long sequence input for better performance. 

\input{tables/model_size}

\input{tables/tab_decompose}
\subsubsection{Model Size Analysis}
Before being introduced into the field of time series prediction, Transformer has shown dominant performance in NLP and CV communities~\cite{vaswani2017attention,kenton2019bert,han2021pre,han2020survey}. One of the key advantages Transformer holds in these fields is being able to increase prediction power through increasing model size. Usually, the model capacity is controlled by Transformer's layer number, which is commonly set between 12 to 128. 
Yet as shown in the experiments of Table \ref{tab:model_size}, when we compare the prediction result with different Transformer models with various numbers of layers, the Transformer with 3 to 6 layers often achieves better results. It raises a question about how to design a proper Transformer architecture with deeper layers to increase the model's capacity and achieve better forecasting performance.

\subsubsection{Seasonal-Trend Decomposition Analysis}

In recent studies, researchers~\cite{xu2021autoformer,zhou2022fedformer,lin2021ssdnet,liu2022pyraformer} begin to realize that the seasonal-trend decomposition~\cite{STL_cleveland1990stl,FastRobustSTL_wen2020} is a crucial part of Transformer's performance in time series forecasting. As an experiment shown in Table \ref{tab:decompose}, we adopt a simple moving average seasonal-trend decomposition architecture proposed in \cite{xu2021autoformer} to test various attention modules. It can be seen that the simple seasonal-trend decomposition model can significantly boost model's performance by 50 \% to 80\%. It is a unique block and such performance boosting through decomposition seems a consistent phenomenon in time series forecasting for Transformer's application, which is worth further investigating for more advanced and carefully designed time series decomposition schemes.

%% file: tables/robustness.tex
\begin{table}[t]
\centering
\caption{The MSE comparisons in robustness experiment of forecasting 96 steps for ETTm2 dataset with prolonging input length.}\vspace{-3mm}
\scalebox{0.75}{
\begin{tabular}{c|c|ccccccccc}
\toprule
\multicolumn{2}{c|}{Model} & Transformer & Autoformer & Informer & Reformer &LogFormer \\
\midrule
\multirow{5}{*}{\rotatebox{90}{$Input\ Len$}}
& 96 & \textbf{0.557} & \textbf{0.239} & 0.428 &\textbf{0.615} &\textbf{0.667}\\
&192 & 0.710 & 0.265 & \textbf{0.385} &0.686 &0.697\\
&336 & 1.078 & 0.375 & 1.078&1.359 &0.937\\
&720 & 1.691 & 0.315 & 1.057&1.443 &2.153\\
&1440& 0.936 & 0.552  &1.898  &0.815 &0.867\\



\bottomrule
\end{tabular}
}
\label{tab:robustness}
\vspace{-3mm}
\end{table}

%% file: tables/model_size.tex
\begin{table}[t]
\centering
\caption{The MSE comparisons in model size experiment of forecasting 96 steps for ETTm2 dataset with different number of layers.}\vspace{-3mm}
\scalebox{0.75}{
\begin{tabular}{c|c|ccccccccc}
\toprule
\multicolumn{2}{c|}{Model} & Transformer & Autoformer & Informer & Reformer &LogFormer \\
\midrule
\multirow{5}{*}{\rotatebox{90}{$Layer \ Num$}}
&3 & 0.557 & \textbf{0.234} & 0.428 &0.597 &0.667\\
&6 &\textbf{0.439}  & 0.282 & 0.489&\textbf{0.353} &\textbf{0.387}\\
&12 &0.556  & 0.238 &0.779&0.481 &0.562\\
&24 &0.580  & 0.266 &0.815&1.109 &0.690\\
&48 &0.461  & NaN  &1.623&OOM &2.992\\



\bottomrule
\end{tabular}
}
\label{tab:model_size}
\vspace{-3mm}
\end{table}

%% file: tables/tab_decompose.tex
\begin{table*}[t!]
\centering
\caption{The MSE comparisons in ablation experiments of seasonal-trend decomposition analysis. 'Ori' means the original version without the decomposition. 'Decomp' means with decomposition. The experiment is performed on ETTm2 dataset with prolonging output length.}\vspace{-3mm}
\scalebox{0.83}{
\begin{tabular}{c|c|cccccccccccccc}
\toprule
\multicolumn{2}{c|}{Model} &
\multicolumn{2}{c|}{FEDformer} &
\multicolumn{2}{c|}{Autoformer} & \multicolumn{2}{c|}{Informer} & \multicolumn{2}{c|}{LogTrans} & \multicolumn{2}{c|}{Reformer} & \multicolumn{2}{c|}{Transformer}  & Promotion \\
\midrule
\multicolumn{2}{c|}{MSE} & Ori & Decomp& Ori & Decomp& Ori & Decomp& Ori & Decomp & Ori & Decomp& Ori & Decomp& Relative&\\
\midrule
\multirow{4}{*}{\rotatebox{90}{$Out\ Len$}}
&96 &0.457& 0.203 & 0.581 & 0.255& 0.365 & 0.354 & 0.768 & 0.231 & 0.658 & 0.218 & 0.604 & 0.204 & 53\%\\
&192&0.841 &0.269 & 1.403 & 0.281& 0.533 & 0.432 &0.989 & 0.378 &1.078 & 0.336 & 1.060 & 0.266 & 62\%\\
&336&1.451 &0.325 & 2.632 & 0.339& 1.363 & 0.481 &1.334 & 0.362 &1.549 & 0.366 &1.413 & 0.375& 75\%\\
&720& 3.282 & 0.421 & 3.058 & 0.422& 3.379 & 0.822 &3.048 & 0.539 &2.631 & 0.502 & 2.672 & 0.537 & 82\%\\



\bottomrule
\end{tabular}
}
\label{tab:decompose}
\vspace{-2mm}
\end{table*}

%% file: 5_2_research_direction.tex
\section{Future Research Opportunities}

Here we highlight a few directions that are potentially promising for future research of Transformers in time series.

\subsection{Inductive Biases for Time Series Transformers}
Vanilla Transformer does not make any assumptions about data patterns and characteristics. Although it is a general and universal network for modeling long-range dependencies, it also comes with a price, i.e., lots of data are needed to train Transformer to improve the generalization and avoid data overfitting.
One of the key features of time series data is its seasonal/periodic and trend patterns~\cite{RobustSTL_wen2018robuststl,STL_cleveland1990stl}. Some recent studies have shown that incorporating series periodicity~\cite{xu2021autoformer} or frequency processing~\cite{zhou2022fedformer} into time series Transformer can enhance performance significantly. Moreover, it is interesting that some studies adopt a seemly opposite inductive bias, but both achieve good numerical improvement: \cite{Nie2022ATS} removes the cross-channel dependency by utilizing a channel-independent attention module, while an interesting work \cite{zhang2023cross} improves its experimental performance by utilizing cross-dimension dependency with a two-stage attention mechanism. Clearly, we have noise and signals in such a cross-channel learning paradigm, but a clever way to utilize such inductive bias to suppress the noise and extract the signal is still desired. 
Thus, one future direction is to consider more effective ways to induce inductive biases into Transformers based on the understanding of time series data and characteristics of specific tasks.








\subsection{Transformers and GNN for Time Series}
Multivariate and spatio-temporal time series are becoming increasingly common in applications, calling for additional techniques to handle high dimensionality, especially the ability to capture the underlying relationships among dimensions. 
Introducing graph neural networks (GNNs) is a natural way to model spatial dependency or relationships among dimensions. Recently, several studies have demonstrated that the combination of GNN and Transformers/attentions could bring not only significant performance improvements like in traffic forecasting~\cite{cai2020traffic,xu2020spatial} and multi-modal forecasting~\cite{GRIN}, but also better understanding of the spatio-temporal dynamics and latent causality. 
It is an important future direction to combine Transformers and GNNs for effectively spatial-temporal modeling in time series.


\subsection{Pre-trained Transformers for Time Series}

Large-scale pre-trained Transformer models have significantly boosted the performance for various tasks in NLP~\cite{kenton2019bert,brown2020language} and CV~\cite{chen2021pre}. 
However, there are limited works on pre-trained Transformers for time series, and existing studies mainly focus on time series classification~\cite{zerveas2021transformer,yang2021voice2series}.
Therefore, how to develop appropriate pre-trained Transformer models for different tasks in time series remains to be examined in the future.

\subsection{Transformers with Architecture Level Variants}
Most developed Transformer models for time series maintain the vanilla Transformer's architecture with modifications mainly in the attention module. We might borrow the idea from Transformer variants in NLP and CV which also have architecture-level model designs to fit different purposes, such as lightweight~\cite{Wu2020LiteTW,Mehta2021DeLighTDA}, cross-block connectivity~\cite{Bapna2018TrainingDN}, adaptive computation time~\cite{dehghani2018universal,Xin2020DeeBERTDE}, and recurrence~\cite{Dai2019TransformerXLAL}. Therefore, one future direction is to consider more architecture-level designs for Transformers specifically optimized for time series data and tasks. 

 
\subsection{Transformers with NAS for Time Series}

Hyper-parameters, such as embedding dimension and the number of heads/layers, can largely affect the performance of Transformers. Manual configuring these hyper-parameters is time-consuming and often results in suboptimal performance. 
AutoML technique like Neural architecture search (NAS)~\cite{NASJMLRSurvey19,MergeNAS} has been a popular technique for discovering effective deep neural architectures, and automating Transformer design using NAS in NLP and CV can be found in recent studies~\cite{so2019evolved,chen2021autoformer}.
For industry-scale time series data which can be of both high dimension and long length, automatically discovering both memory- and computational-efficient Transformer architectures is of practical importance, making it an important future direction for time series Transformers.

%% file: 6_conclusion.tex
\section{Conclusion}
We have provided a survey on time series Transformers. We organize the reviewed methods in a new taxonomy consisting of network design and application. We summarize representative methods in each category, discuss their strengths and limitations by experimental evaluation, and highlight future research directions.

%% file: 0_main_TSformer.bbl
\begin{thebibliography}{}

\bibitem[\protect\citeauthoryear{Bapna \bgroup \em et al.\egroup
  }{2018}]{Bapna2018TrainingDN}
Ankur Bapna, Mia~Xu Chen, Orhan Firat, Yuan Cao, and Yonghui Wu.
\newblock Training deeper neural machine translation models with transparent
  attention.
\newblock In {\em EMNLP}, 2018.

\bibitem[\protect\citeauthoryear{Benidis \bgroup \em et al.\egroup
  }{2022}]{benidis2020neural}
Konstantinos Benidis, Syama~Sundar Rangapuram, Valentin Flunkert, Yuyang Wang,
  Danielle Maddix, , et~al.
\newblock Deep learning for time series forecasting: Tutorial and literature
  survey.
\newblock {\em ACM Computing Surveys}, 55(6):1--36, 2022.

\bibitem[\protect\citeauthoryear{Bl{\'a}zquez-Garc{\'\i}a \bgroup \em et
  al.\egroup }{2021}]{blazquez2021review}
Ane Bl{\'a}zquez-Garc{\'\i}a, Angel Conde, Usue Mori, et~al.
\newblock A review on outlier/anomaly detection in time series data.
\newblock {\em ACM Computing Surveys}, 54(3):1--33, 2021.

\bibitem[\protect\citeauthoryear{Brown \bgroup \em et al.\egroup
  }{2020}]{brown2020language}
Tom Brown, Benjamin Mann, Nick Ryder, Melanie Subbiah, Jared~D Kaplan, Prafulla
  Dhariwal, et~al.
\newblock Language models are few-shot learners.
\newblock {\em NeurIPS}, 2020.

\bibitem[\protect\citeauthoryear{Cai \bgroup \em et al.\egroup
  }{2020}]{cai2020traffic}
Ling Cai, Krzysztof Janowicz, Gengchen Mai, Bo~Yan, and Rui Zhu.
\newblock Traffic transformer: Capturing the continuity and periodicity of time
  series for traffic forecasting.
\newblock {\em Transactions in GIS}, 24(3):736--755, 2020.

\bibitem[\protect\citeauthoryear{Chen \bgroup \em et al.\egroup
  }{2021a}]{chen2021pre}
Hanting Chen, Yunhe Wang, Tianyu Guo, Chang Xu, Yiping Deng, Zhenhua Liu, Siwei
  Ma, Chunjing Xu, et~al.
\newblock Pre-trained image processing transformer.
\newblock In {\em CVPR}, 2021.

\bibitem[\protect\citeauthoryear{Chen \bgroup \em et al.\egroup
  }{2021b}]{chen2021autoformer}
Minghao Chen, Houwen Peng, Jianlong Fu, and Haibin Ling.
\newblock {AutoFormer}: Searching transformers for visual recognition.
\newblock In {\em CVPR}, 2021.

\bibitem[\protect\citeauthoryear{Chen \bgroup \em et al.\egroup
  }{2021c}]{chen2021learning}
Zekai Chen, Dingshuo Chen, Xiao Zhang, Zixuan Yuan, and Xiuzhen Cheng.
\newblock Learning graph structures with transformer for multivariate time
  series anomaly detection in {IoT}.
\newblock {\em IEEE Internet of Things Journal}, 2021.

\bibitem[\protect\citeauthoryear{Chen \bgroup \em et al.\egroup
  }{2022}]{chen2022quatformer}
Weiqi Chen, Wenwei Wang, Bingqing Peng, Qingsong Wen, Tian Zhou, and Liang Sun.
\newblock Learning to rotate: Quaternion transformer for complicated periodical
  time series forecasting.
\newblock In {\em KDD}, 2022.

\bibitem[\protect\citeauthoryear{Choi \bgroup \em et al.\egroup
  }{2021}]{choi2021deep}
Kukjin Choi, Jihun Yi, Changhwa Park, and Sungroh Yoon.
\newblock Deep learning for anomaly detection in time-series data: Review,
  analysis, and guidelines.
\newblock {\em IEEE Access}, 2021.

\bibitem[\protect\citeauthoryear{Chowdhury \bgroup \em et al.\egroup
  }{2022}]{chowdhury2022tarnet}
Ranak~Roy Chowdhury, Xiyuan Zhang, Jingbo Shang, Rajesh~K Gupta, and Dezhi
  Hong.
\newblock {TARNet}: Task-aware reconstruction for time-series transformer.
\newblock In {\em KDD}, 2022.

\bibitem[\protect\citeauthoryear{Cirstea \bgroup \em et al.\egroup
  }{2022}]{cirstea2022triformer}
Razvan-Gabriel Cirstea, Chenjuan Guo, Bin Yang, Tung Kieu, Xuanyi Dong, and
  Shirui Pan.
\newblock Triformer: Triangular, variable-specific attentions for long sequence
  multivariate time series forecasting.
\newblock In {\em IJCAI}, 2022.

\bibitem[\protect\citeauthoryear{Cleveland \bgroup \em et al.\egroup
  }{1990}]{STL_cleveland1990stl}
Robert Cleveland, William Cleveland, Jean McRae, et~al.
\newblock {STL}: A seasonal-trend decomposition procedure based on loess.
\newblock {\em Journal of Official Statistics}, 6(1):3--73, 1990.

\bibitem[\protect\citeauthoryear{Dai \bgroup \em et al.\egroup
  }{2019}]{Dai2019TransformerXLAL}
Zihang Dai, Zhilin Yang, Yiming Yang, Jaime~G. Carbonell, Quoc~V. Le, et~al.
\newblock {Transformer-XL}: Attentive language models beyond a fixed-length
  context.
\newblock In {\em ACL}, 2019.

\bibitem[\protect\citeauthoryear{Dehghani \bgroup \em et al.\egroup
  }{2019}]{dehghani2018universal}
Mostafa Dehghani, Stephan Gouws, Oriol Vinyals, Jakob Uszkoreit, and {\L}ukasz
  Kaiser.
\newblock Universal transformers.
\newblock In {\em ICLR}, 2019.

\bibitem[\protect\citeauthoryear{Dong \bgroup \em et al.\egroup
  }{2018}]{dong2018speech}
Linhao Dong, Shuang Xu, and Bo~Xu.
\newblock Speech-transformer: a no-recurrence sequence-to-sequence model for
  speech recognition.
\newblock In {\em ICASSP}, 2018.

\bibitem[\protect\citeauthoryear{Dosovitskiy \bgroup \em et al.\egroup
  }{2021}]{Dosovitskiy2020AnII}
Alexey Dosovitskiy, Lucas Beyer, Alexander Kolesnikov, Dirk Weissenborn,
  Xiaohua Zhai, Thomas Unterthiner, et~al.
\newblock An image is worth 16x16 words: Transformers for image recognition at
  scale.
\newblock In {\em ICLR}, 2021.

\bibitem[\protect\citeauthoryear{Elsken \bgroup \em et al.\egroup
  }{2019}]{NASJMLRSurvey19}
Elsken, Thomas, Jan~Hendrik Metzen, and Frank Hutter.
\newblock Neural architecture search: A survey.
\newblock {\em Journal of Machine Learning Research}, 2019.

\bibitem[\protect\citeauthoryear{Gao \bgroup \em et al.\egroup
  }{2022}]{gaoearthformer2022}
Zhihan Gao, Xingjian Shi, Hao Wang, Yi~Zhu, Bernie Wang, Mu~Li, et~al.
\newblock Earthformer: Exploring space-time transformers for earth system
  forecasting.
\newblock In {\em NeurIPS}, 2022.

\bibitem[\protect\citeauthoryear{Gehring \bgroup \em et al.\egroup
  }{2017}]{gehring2017convolutional}
Jonas Gehring, Michael Auli, David Grangier, Denis Yarats, and Yann~N Dauphin.
\newblock Convolutional sequence to sequence learning.
\newblock In {\em ICML}, 2017.

\bibitem[\protect\citeauthoryear{Goodfellow \bgroup \em et al.\egroup
  }{2014}]{goodfellow2014generative}
Ian Goodfellow, Jean Pouget-Abadie, Mehdi Mirza, Bing Xu, David Warde-Farley,
  Sherjil Ozair, et~al.
\newblock Generative adversarial nets.
\newblock {\em NeurIPS}, 2014.

\bibitem[\protect\citeauthoryear{Han \bgroup \em et al.\egroup
  }{2021}]{han2021pre}
Xu~Han, Zhengyan Zhang, Ning Ding, Yuxian Gu, Xiao Liu, Yuqi Huo, Jiezhong Qiu,
  Liang Zhang, et~al.
\newblock Pre-trained models: Past, present and future.
\newblock {\em AI Open}, 2021.

\bibitem[\protect\citeauthoryear{Han \bgroup \em et al.\egroup
  }{2022}]{han2020survey}
Kai Han, Yunhe Wang, Hanting Chen, Xinghao Chen, Jianyuan Guo, Zhenhua Liu,
  Yehui Tang, An~Xiao, et~al.
\newblock A survey on vision transformer.
\newblock {\em IEEE TPAMI}, 45(1):87--110, 2022.

\bibitem[\protect\citeauthoryear{Hyndman and
  Khandakar}{2008}]{hyndman2008automatic}
Rob~J Hyndman and Yeasmin Khandakar.
\newblock Automatic time series forecasting: the forecast package for r.
\newblock {\em Journal of statistical software}, 27:1--22, 2008.

\bibitem[\protect\citeauthoryear{Ismail~Fawaz \bgroup \em et al.\egroup
  }{2019}]{ismail2019deep}
Hassan Ismail~Fawaz, Germain Forestier, Jonathan Weber, Lhassane Idoumghar, and
  Pierre-Alain Muller.
\newblock Deep learning for time series classification: a review.
\newblock {\em Data mining and knowledge discovery}, 2019.

\bibitem[\protect\citeauthoryear{Ke \bgroup \em et al.\egroup
  }{2021}]{ke2020rethinking}
Guolin Ke, Di~He, and Tie-Yan Liu.
\newblock Rethinking positional encoding in language pre-training.
\newblock In {\em ICLR}, 2021.

\bibitem[\protect\citeauthoryear{Kenton and others}{2019}]{kenton2019bert}
Jacob Devlin Ming-Wei~Chang Kenton et~al.
\newblock {BERT}: Pre-training of deep bidirectional transformers for language
  understanding.
\newblock In {\em NAACL-HLT}, 2019.

\bibitem[\protect\citeauthoryear{Kingma and Welling}{2014}]{kingma2013auto}
Diederik~P Kingma and Max Welling.
\newblock Auto-encoding variational bayes.
\newblock In {\em ICLR}, 2014.

\bibitem[\protect\citeauthoryear{Li \bgroup \em et al.\egroup
  }{2019}]{li2019enhancing}
Shiyang Li, Xiaoyong Jin, Yao Xuan, Xiyou Zhou, Wenhu Chen, Yu-Xiang Wang, and
  Xifeng Yan.
\newblock Enhancing the locality and breaking the memory bottleneck of
  transformer on time series forecasting.
\newblock In {\em NeurIPS}, 2019.

\bibitem[\protect\citeauthoryear{Li \bgroup \em et al.\egroup }{2021}]{GRIN}
Longyuan Li, Jian Yao, Li~Wenliang, Tong He, Tianjun Xiao, Junchi Yan, David
  Wipf, and Zheng Zhang.
\newblock Grin: Generative relation and intention network for multi-agent
  trajectory prediction.
\newblock In {\em NeurIPS}, 2021.

\bibitem[\protect\citeauthoryear{Liang \bgroup \em et al.\egroup
  }{2023}]{liang2023airformer}
Yuxuan Liang, Yutong Xia, Songyu Ke, Yiwei Wang, Qingsong Wen, Junbo Zhang,
  Yu~Zheng, and Roger Zimmermann.
\newblock {AirFormer}: Predicting nationwide air quality in china with
  transformers.
\newblock In {\em AAAI}, 2023.

\bibitem[\protect\citeauthoryear{Lim and Zohren}{2021}]{tsDLSurvey21}
Bryan Lim and Stefan Zohren.
\newblock Time-series forecasting with deep learning: a survey.
\newblock {\em Philosophical Transactions of the Royal Society}, 2021.

\bibitem[\protect\citeauthoryear{Lim \bgroup \em et al.\egroup
  }{2021}]{lim2021temporal}
Bryan Lim, Sercan~{\"O} Ar{\i}k, Nicolas Loeff, and Tomas Pfister.
\newblock Temporal fusion transformers for interpretable multi-horizon time
  series forecasting.
\newblock {\em International Journal of Forecasting}, 37(4):1748--1764, 2021.

\bibitem[\protect\citeauthoryear{Lin \bgroup \em et al.\egroup
  }{2021}]{lin2021ssdnet}
Yang Lin, Irena Koprinska, and Mashud Rana.
\newblock {SSDNet}: State space decomposition neural network for time series
  forecasting.
\newblock In {\em ICDM}, 2021.

\bibitem[\protect\citeauthoryear{Liu \bgroup \em et al.\egroup
  }{2021}]{liu2021gated}
Minghao Liu, Shengqi Ren, Siyuan Ma, Jiahui Jiao, Yizhou Chen, Zhiguang Wang,
  and Wei Song.
\newblock Gated transformer networks for multivariate time series
  classification.
\newblock {\em arXiv preprint arXiv:2103.14438}, 2021.

\bibitem[\protect\citeauthoryear{Liu \bgroup \em et al.\egroup
  }{2022a}]{liu2022pyraformer}
Shizhan Liu, Hang Yu, Cong Liao, Jianguo Li, Weiyao Lin, Alex~X. Liu, and
  Schahram Dustdar.
\newblock Pyraformer: Low-complexity pyramidal attention for long-range time
  series modeling and forecasting.
\newblock In {\em ICLR}, 2022.

\bibitem[\protect\citeauthoryear{Liu \bgroup \em et al.\egroup
  }{2022b}]{liu2022non}
Yong Liu, Haixu Wu, Jianmin Wang, and Mingsheng Long.
\newblock Non-stationary transformers: Exploring the stationarity in time
  series forecasting.
\newblock In {\em NeurIPS}, 2022.

\bibitem[\protect\citeauthoryear{Mehta \bgroup \em et al.\egroup
  }{2021}]{Mehta2021DeLighTDA}
Sachin Mehta, Marjan Ghazvininejad, Srini Iyer, Luke Zettlemoyer, and Hannaneh
  Hajishirzi.
\newblock Delight: Deep and light-weight transformer.
\newblock In {\em ICLR}, 2021.

\bibitem[\protect\citeauthoryear{Mei \bgroup \em et al.\egroup
  }{2022}]{mei2022transformer}
Hongyuan Mei, Chenghao Yang, and Jason Eisner.
\newblock Transformer embeddings of irregularly spaced events and their
  participants.
\newblock In {\em ICLR}, 2022.

\bibitem[\protect\citeauthoryear{Nie \bgroup \em et al.\egroup
  }{2023}]{Nie2022ATS}
Yuqi Nie, Nam~H. Nguyen, Phanwadee Sinthong, and Jayant Kalagnanam.
\newblock A time series is worth 64 words: Long-term forecasting with
  transformers.
\newblock In {\em ICLR}, 2023.

\bibitem[\protect\citeauthoryear{Rußwurm and
  Körner}{2020}]{attention-satellite-classi/rubwurm/2020}
Marc Rußwurm and Marco Körner.
\newblock Self-attention for raw optical satellite time series classification.
\newblock {\em ISPRS J. Photogramm. Remote Sens.}, 169:421--435, 11 2020.

\bibitem[\protect\citeauthoryear{Shabani \bgroup \em et al.\egroup
  }{2023}]{shabani2023scaleformer}
Amin Shabani, Amir Abdi, Lili Meng, and Tristan Sylvain.
\newblock Scaleformer: iterative multi-scale refining transformers for time
  series forecasting.
\newblock In {\em ICLR}, 2023.

\bibitem[\protect\citeauthoryear{Shaw \bgroup \em et al.\egroup
  }{2018}]{shaw2018self}
Peter Shaw, Jakob Uszkoreit, and Ashish Vaswani.
\newblock Self-attention with relative position representations.
\newblock In {\em NAACL}, 2018.

\bibitem[\protect\citeauthoryear{Shchur \bgroup \em et al.\egroup
  }{2021}]{shchur2021neural}
Oleksandr Shchur, Ali~Caner T{\"u}rkmen, Tim Januschowski, and Stephan
  G{\"u}nnemann.
\newblock Neural temporal point processes: A review.
\newblock In {\em IJCAI}, 2021.

\bibitem[\protect\citeauthoryear{So \bgroup \em et al.\egroup
  }{2019}]{so2019evolved}
David So, Quoc Le, and Chen Liang.
\newblock The evolved transformer.
\newblock In {\em ICML}, 2019.

\bibitem[\protect\citeauthoryear{Tang and
  Matteson}{2021}]{tang2021probabilistic}
Binh Tang and David Matteson.
\newblock Probabilistic transformer for time series analysis.
\newblock In {\em NeurIPS}, 2021.

\bibitem[\protect\citeauthoryear{Tay \bgroup \em et al.\egroup
  }{2022}]{tay2020efficient}
Yi~Tay, Mostafa Dehghani, Dara Bahri, and Donald Metzler.
\newblock Efficient transformers: A survey.
\newblock {\em ACM Computing Surveys}, 55(6):1--28, 2022.

\bibitem[\protect\citeauthoryear{Torres \bgroup \em et al.\egroup
  }{2021}]{DLTSSurvey21}
José~F. Torres, Dalil Hadjout, Abderrazak Sebaa, Francisco Martínez-Álvarez,
  and Alicia Troncoso.
\newblock Deep learning for time series forecasting: a survey.
\newblock {\em Big Data}, 2021.

\bibitem[\protect\citeauthoryear{Tuli \bgroup \em et al.\egroup
  }{2022}]{tuli2022tranad}
Shreshth Tuli, Giuliano Casale, and Nicholas~R Jennings.
\newblock {TranAD}: Deep transformer networks for anomaly detection in
  multivariate time series data.
\newblock In {\em VLDB}, 2022.

\bibitem[\protect\citeauthoryear{Vaswani \bgroup \em et al.\egroup
  }{2017}]{vaswani2017attention}
Ashish Vaswani, Noam Shazeer, Niki Parmar, Jakob Uszkoreit, Llion Jones,
  Aidan~N Gomez, {\L}ukasz Kaiser, et~al.
\newblock Attention is all you need.
\newblock In {\em NeurIPS}, 2017.

\bibitem[\protect\citeauthoryear{Wang \bgroup \em et al.\egroup
  }{2020}]{MergeNAS}
Xiaoxing Wang, Chao Xue, Junchi Yan, Xiaokang Yang, Yonggang Hu, et~al.
\newblock {MergeNAS}: Merge operations into one for differentiable architecture
  search.
\newblock In {\em IJCAI}, 2020.

\bibitem[\protect\citeauthoryear{Wang \bgroup \em et al.\egroup
  }{2022}]{wang2022variational}
Xixuan Wang, Dechang Pi, Xiangyan Zhang, et~al.
\newblock Variational transformer-based anomaly detection approach for
  multivariate time series.
\newblock {\em Measurement}, page 110791, 2022.

\bibitem[\protect\citeauthoryear{Wen \bgroup \em et al.\egroup
  }{2019}]{RobustSTL_wen2018robuststl}
Qingsong Wen, Jingkun Gao, Xiaomin Song, Liang Sun, Huan Xu, et~al.
\newblock {RobustSTL}: A robust seasonal-trend decomposition algorithm for long
  time series.
\newblock In {\em AAAI}, 2019.

\bibitem[\protect\citeauthoryear{Wen \bgroup \em et al.\egroup
  }{2020}]{FastRobustSTL_wen2020}
Qingsong Wen, Zhe Zhang, Yan Li, and Liang Sun.
\newblock {Fast RobustSTL}: Efficient and robust seasonal-trend decomposition
  for time series with complex patterns.
\newblock In {\em KDD}, 2020.

\bibitem[\protect\citeauthoryear{Wen \bgroup \em et al.\egroup
  }{2021a}]{WenRobustPeriod20}
Qingsong Wen, Kai He, Liang Sun, Yingying Zhang, Min Ke, et~al.
\newblock {RobustPeriod}: Time-frequency mining for robust multiple
  periodicities detection.
\newblock In {\em SIGMOD}, 2021.

\bibitem[\protect\citeauthoryear{Wen \bgroup \em et al.\egroup
  }{2021b}]{wentsaug2021}
Qingsong Wen, Liang Sun, Fan Yang, Xiaomin Song, Jingkun Gao, Xue Wang, and
  Huan Xu.
\newblock Time series data augmentation for deep learning: A survey.
\newblock In {\em IJCAI}, 2021.

\bibitem[\protect\citeauthoryear{Wen \bgroup \em et al.\egroup
  }{2022}]{wen2022robust}
Qingsong Wen, Linxiao Yang, Tian Zhou, and Liang Sun.
\newblock Robust time series analysis and applications: An industrial
  perspective.
\newblock In {\em KDD}, 2022.

\bibitem[\protect\citeauthoryear{Wu \bgroup \em et al.\egroup
  }{2020a}]{wu2020adversarial}
Sifan Wu, Xi~Xiao, Qianggang Ding, Peilin Zhao, Ying Wei, and Junzhou Huang.
\newblock Adversarial sparse transformer for time series forecasting.
\newblock In {\em NeurIPS}, 2020.

\bibitem[\protect\citeauthoryear{Wu \bgroup \em et al.\egroup
  }{2020b}]{Wu2020LiteTW}
Zhanghao Wu, Zhijian Liu, Ji~Lin, Yujun Lin, and Song Han.
\newblock Lite transformer with long-short range attention.
\newblock In {\em ICLR}, 2020.

\bibitem[\protect\citeauthoryear{Wu \bgroup \em et al.\egroup
  }{2021}]{xu2021autoformer}
Haixu Wu, Jiehui Xu, Jianmin Wang, and Mingsheng Long.
\newblock Autoformer: Decomposition transformers with auto-correlation for
  long-term series forecasting.
\newblock In {\em NeurIPS}, 2021.

\bibitem[\protect\citeauthoryear{Xin \bgroup \em et al.\egroup
  }{2020}]{Xin2020DeeBERTDE}
Ji~Xin, Raphael Tang, Jaejun Lee, Yaoliang Yu, and Jimmy~J. Lin.
\newblock {DeeBERT}: Dynamic early exiting for accelerating bert inference.
\newblock In {\em ACL}, 2020.

\bibitem[\protect\citeauthoryear{Xu \bgroup \em et al.\egroup
  }{2020}]{xu2020spatial}
Mingxing Xu, Wenrui Dai, Chunmiao Liu, Xing Gao, Weiyao Lin, Guo-Jun Qi, and
  Hongkai Xiong.
\newblock Spatial-temporal transformer networks for traffic flow forecasting.
\newblock {\em arXiv preprint arXiv:2001.02908}, 2020.

\bibitem[\protect\citeauthoryear{Xu \bgroup \em et al.\egroup
  }{2022}]{xu2022anomalyTrans}
Jiehui Xu, Haixu Wu, Jianmin Wang, and Mingsheng Long.
\newblock {Anomaly Transformer}: Time series anomaly detection with association
  discrepancy.
\newblock In {\em ICLR}, 2022.

\bibitem[\protect\citeauthoryear{Yan \bgroup \em et al.\egroup
  }{2019}]{yan2019modeling}
Junchi Yan, Hongteng Xu, and Liangda Li.
\newblock Modeling and applications for temporal point processes.
\newblock In {\em KDD}, 2019.

\bibitem[\protect\citeauthoryear{Yang \bgroup \em et al.\egroup
  }{2021}]{yang2021voice2series}
Chao-Han~Huck Yang, Yun-Yun Tsai, and Pin-Yu Chen.
\newblock Voice2series: Reprogramming acoustic models for time series
  classification.
\newblock In {\em ICML}, 2021.

\bibitem[\protect\citeauthoryear{Yu \bgroup \em et al.\egroup
  }{2020}]{yu2020spatio}
Cunjun Yu, Xiao Ma, Jiawei Ren, Haiyu Zhao, and Shuai Yi.
\newblock Spatio-temporal graph transformer networks for pedestrian trajectory
  prediction.
\newblock In {\em ECCV}, 2020.

\bibitem[\protect\citeauthoryear{Yuan and Lin}{2020}]{yuan2020self}
Yuan Yuan and Lei Lin.
\newblock Self-supervised pretraining of transformers for satellite image time
  series classification.
\newblock {\em IEEE J-STARS}, 14:474--487, 2020.

\bibitem[\protect\citeauthoryear{Yun \bgroup \em et al.\egroup
  }{2020}]{Yun2019AreTU}
Chulhee Yun, Srinadh Bhojanapalli, Ankit~Singh Rawat, Sashank~J. Reddi, et~al.
\newblock Are transformers universal approximators of sequence-to-sequence
  functions?
\newblock In {\em ICLR}, 2020.

\bibitem[\protect\citeauthoryear{Zeng \bgroup \em et al.\egroup
  }{2023}]{zeng2022transformers}
Ailing Zeng, Muxi Chen, Lei Zhang, and Qiang Xu.
\newblock Are transformers effective for time series forecasting?
\newblock In {\em AAAI}, 2023.

\bibitem[\protect\citeauthoryear{Zerveas \bgroup \em et al.\egroup
  }{2021}]{zerveas2021transformer}
George Zerveas, Srideepika Jayaraman, Dhaval Patel, Anuradha Bhamidipaty, and
  Carsten Eickhoff.
\newblock A transformer-based framework for multivariate time series
  representation learning.
\newblock In {\em KDD}, 2021.

\bibitem[\protect\citeauthoryear{Zhang and Yan}{2023}]{zhang2023cross}
Yunhao Zhang and Junchi Yan.
\newblock Crossformer: Transformer utilizing cross-dimension dependency for
  multivariate time series forecasting.
\newblock In {\em ICLR}, 2023.

\bibitem[\protect\citeauthoryear{Zhang \bgroup \em et al.\egroup
  }{2020}]{zhang2020self}
Qiang Zhang, Aldo Lipani, Omer Kirnap, and Emine Yilmaz.
\newblock Self-attentive {Hawkes} process.
\newblock In {\em ICML}, 2020.

\bibitem[\protect\citeauthoryear{Zhang \bgroup \em et al.\egroup
  }{2021}]{zhang2021unsupervised}
Hongwei Zhang, Yuanqing Xia, et~al.
\newblock Unsupervised anomaly detection in multivariate time series through
  transformer-based variational autoencoder.
\newblock In {\em CCDC}, 2021.

\bibitem[\protect\citeauthoryear{Zhou \bgroup \em et al.\egroup
  }{2021}]{zhou2021informer}
Haoyi Zhou, Shanghang Zhang, Jieqi Peng, Shuai Zhang, Jianxin Li, Hui Xiong,
  and Wancai Zhang.
\newblock Informer: Beyond efficient transformer for long sequence time-series
  forecasting.
\newblock In {\em AAAI}, 2021.

\bibitem[\protect\citeauthoryear{Zhou \bgroup \em et al.\egroup
  }{2022}]{zhou2022fedformer}
Tian Zhou, Ziqing Ma, Qingsong Wen, Xue Wang, Liang Sun, and Rong Jin.
\newblock {FEDformer}: Frequency enhanced decomposed transformer for long-term
  series forecasting.
\newblock In {\em ICML}, 2022.

\bibitem[\protect\citeauthoryear{Zuo \bgroup \em et al.\egroup
  }{2020}]{zuo2020transformer}
Simiao Zuo, Haoming Jiang, Zichong Li, Tuo Zhao, and Hongyuan Zha.
\newblock Transformer {Hawkes} process.
\newblock In {\em ICML}, 2020.

\end{thebibliography}
